%% file: main.tex
\documentclass[10pt,twocolumn,letterpaper]{article}
\usepackage[pagenumbers]{cvpr} %

\input{macro}

\usepackage{microtype}
\definecolor{cvprblue}{rgb}{0.21,0.49,0.74}
\usepackage[pagebackref,breaklinks,colorlinks,citecolor=cvprblue,linkcolor=cornellred]{hyperref}

\title{FiCA: Feed-forward Instant Gaussian Codec Avatars\\from a Single Portrait Image\vspace{-3mm}}

\def\authorBlock{
    Kim Youwang${}^{1,2*}$ \quad
    Zhengyu Yang${}^{1}$ \quad
    Liuhao Ge${}^{1}$ \quad
    Yu Rong${}^{1}$ \quad
    Timur Bagautdinov${}^{1}$ \quad
    Su Zhaoen${}^{1}$\\
    Nir Sopher${}^{1}$ \quad
    Jovan Popovi\'c${}^{1}$ \quad
    Teng Deng${}^{1}$ \quad
    Tae-Hyun Oh${}^{2,3}$ \quad
    Chen Cao${}^{1}$ \vspace{1.5mm}\\
   \small{${}^{1}$Codec Avatars Lab, Meta\quad ${}^{2}$Dept. of Electrical Engineering, POSTECH\quad ${}^{3}$School of Computing, KAIST}\\ 
   \small{\url{https://kim-youwang.github.io/FiCA}}
   \vspace{-1.5mm}
}
\author{\authorBlock}

\begin{document}
\twocolumn[{%
\renewcommand\twocolumn[1][]{#1}%
\maketitle 
\vspace{-3mm}
\input{figures/tex/teaser}
}]

{
  \renewcommand{\thefootnote}%
    {\fnsymbol{footnote}}
  \footnotetext[1]{Work done while Youwang was an intern at Codec Avatars Lab, Meta.}
}

\input{sec/0_abstract}    
\vspace{-3mm}
\input{sec/1_intro}

\input{sec/2_related}

\input{sec/3_method}

\input{sec/4_exp}

\input{sec/5_conc}

{
    \small
    \bibliographystyle{ieeenat_fullname}
    \bibliography{main}
}

\input{supp}

\end{document}

%% file: macro.tex
\usepackage{enumitem}
\usepackage{multirow}
\usepackage{wrapfig}
\usepackage{colortbl}
\usepackage[normalem]{ulem}
\usepackage{microtype}
\usepackage{comment}

\setlist[itemize]{align=parleft,left=0pt,topsep=1mm,itemsep=0mm,parsep=1mm}

\definecolor{azure(colorwheel)}{rgb}{0.0, 0.5, 1.0}
\definecolor{nicegreen}{rgb}{0.0, 0.7, 0.1}
\definecolor{yw}{rgb}{0.01176, 0.5490, 0.5490}
\definecolor{ashblue}{rgb}{0.36, 0.54, 0.66}
\definecolor{ashgrey}{rgb}{0.7, 0.75, 0.71}
\definecolor{applegreen}{rgb}{0.55, 0.71, 0.0}
\definecolor{blue}{rgb}{0.0, 0.0, 1.0}
\definecolor{postechred}{rgb}{0.784, 0.003, 0.313}
\definecolor{ywg}{rgb}{0.9960, 0.8984, 0.5859}
\definecolor{ballblue}{rgb}{0.13, 0.67, 0.8}
\definecolor{cornellred}{rgb}{0.7, 0.11, 0.11}
\definecolor{darkcyan}{rgb}{0.0, 0.55, 0.55}
\definecolor{CuGray}{gray}{0.9}
\definecolor{airforceblue}{rgb}{0.36, 0.54, 0.66}
\definecolor{rev}{rgb}{0.784, 0.003, 0.313}
\definecolor{pink}{cmyk}{0, 0.7808, 0.4429, 0.1412}
\definecolor{amethyst}{rgb}{0.6, 0.4, 0.8}
\definecolor{black}{rgb}{0.0, 0.0, 0.0}
\definecolor{tb3_yellow}{rgb}{0.996, 1.0, 0.6}
\definecolor{tb3_orange}{rgb}{0.980, 0.8, 0.604}
\definecolor{tb3_red}{rgb}{0.972, 0.6, 0.6}
\definecolor{dimgray}{rgb}{0.41, 0.41, 0.41}
\definecolor{brickred}{rgb}{0.8, 0.25, 0.33}
\definecolor{bleudefrance}{rgb}{0.19, 0.55, 0.91}
\definecolor{blue(ncs)}{rgb}{0.265, 0.445, 0.765}
\definecolor{blue(ryb)}{rgb}{0.01, 0.28, 1.0}
\definecolor{orange}{rgb}{1.0, 0.49, 0.0}
\definecolor{Gray}{gray}{0.88}
\definecolor{green(ncs)}{rgb}{0.0, 0.62, 0.42}
\definecolor{brightpink}{rgb}{1.0, 0.0, 0.5}
\definecolor{pastelred}{rgb}{0.66, 0.25, 0.28}
\definecolor{pastelorange}{rgb}{0.54, 0.38, 0.30}
\definecolor{pastelgreen}{rgb}{0.39, 0.55, 0.38}
\definecolor{pastelblue}{rgb}{0.34, 0.42, 0.90}
\definecolor{pastelpurple}{rgb}{0.50, 0.30, 0.50}
\definecolor{c_diff}{rgb}{0.3058823529, 0.5843137255, 0.8509803922}
\definecolor{c_urn}{rgb}{0.98, 0.4941176471, 0.4745098039}
\definecolor{c_upm}{rgb}{0.3058823529, 0.6549019608, 0.1803921569}
\definecolor{c_exp}{rgb}{1.0, 0.7529411765, 0.0}
\definecolor{iccvblue}{rgb}{0.21,0.49,0.74}

\definecolor{kellygreen}{rgb}{0.3, 0.73, 0.09}
\newcommand{\colorref}[1]{{\color{cornellred}{#1}}}

\newcolumntype{g}{>{\columncolor{CuGray}}c}
\newcolumntype{z}{>{\columncolor{CuGray}}l}

\renewcommand{\paragraph}[1]{\vspace{1mm}\noindent\textbf{#1.}\,\,}

\definecolor{tabfirst}{rgb}{1, 0.7, 0.7} %
\definecolor{tabsecond}{rgb}{1, 0.85, 0.7} %
\definecolor{tabthird}{rgb}{1, 1, 0.7} %

\def\ours{FiCA}

\usepackage{xspace}

\makeatletter
\def\@fnsymbol#1{\ensuremath{\ifcase#1\or *\or \dagger\or \ddagger\or
   \mathsection\or \mathparagraph\or \|\or **\or \dagger\dagger
   \or \ddagger\ddagger \else\@ctrerr\fi}}
\makeatother

\def\onedot{.\@\xspace}
\def\eg{\emph{e.g}\onedot} 
\def\ie{\emph{i.e}\onedot}

\def\etal{\emph{et al}\onedot}

\newcommand{\Sref}[1]{Sec.~\ref{#1}}
\newcommand{\Eref}[1]{Eq.~(\ref{#1})}
\newcommand{\Fref}[1]{Fig.~\ref{#1}}
\newcommand{\Tref}[1]{Table~\ref{#1}}

\newcommand{\bc}{{\mathbf{c}}}
\newcommand{\bd}{{\mathbf{d}}}

\newcommand{\bk}{{\mathbf{k}}}

\newcommand{\bm}{{\mathbf{m}}}

\newcommand{\bo}{{\mathbf{o}}}

\newcommand{\bq}{{\mathbf{q}}}

\newcommand{\bs}{{\mathbf{s}}}

\newcommand{\bv}{{\mathbf{v}}}

\newcommand{\bx}{{\mathbf{x}}}

\newcommand{\bz}{{\mathbf{z}}}

\newcommand{\bG}{\mathbf{G}}

\newcommand{\bI}{\mathbf{I}}

\newcommand{\bL}{\mathbf{L}}

\newcommand{\bN}{\mathbf{N}}

\newcommand{\bT}{\mathbf{T}}

\newcommand{\be}{\begin{eqnarray}}
\newcommand{\ee}{\end{eqnarray}}
\newcommand{\bee}{\begin{eqnarray*}}
\newcommand{\eee}{\end{eqnarray*}}

\newcommand{\matrixb}{\left[ \begin{array}}
\newcommand{\matrixe}{\end{array} \right]}

%% file: figures/tex/teaser.tex
\begin{center}
\vspace{-6mm}
  \centering
  \captionsetup{type=figure}
  \includegraphics[width=\linewidth]{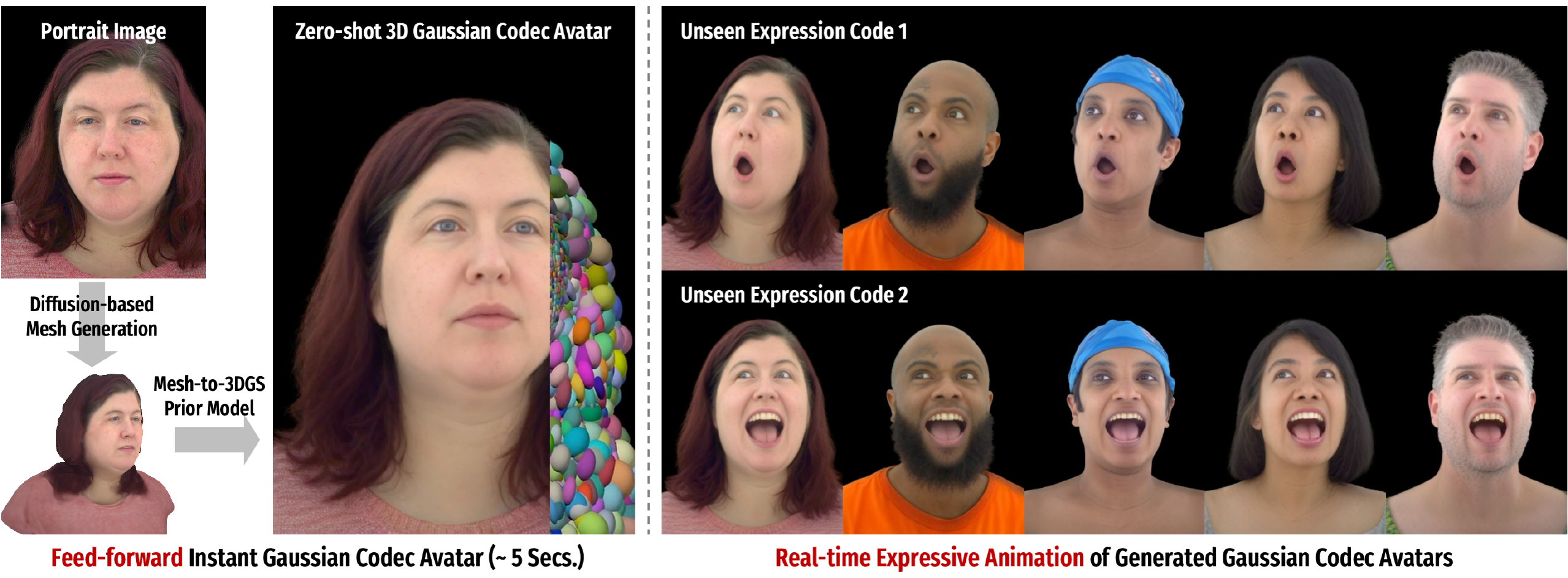}
   \captionof{figure}{\textbf{Feed-forward instant Gaussian Codec Avatars (FiCA).} Our method creates drivable, photorealistic 3D Gaussian head avatars from a casually captured, single portrait image, within \emph{5 seconds}. 
   The generated head avatars can be animated consistently across different identities in real-time, given target expressions.
   Please refer to the supplementary video for dynamic avatar animation results.
   }
   \label{fig:teaser}
\end{center}

%% file: sec/0_abstract.tex
\begin{abstract}
We introduce \textbf{\ours}, a \textbf{F}eed-forward, \textbf{i}nstant Gaussian \textbf{C}odec \textbf{A}vatar generation pipeline that creates lifelike avatars from a single portrait image.
Generating a photorealistic and drivable avatar from just a single image is significantly challenging due to the limited visual information available to accurately infer the 3D appearance and geometry of human heads.
To address this, we develop a novel system that combines human-centric vision foundation models with a diffusion model. This system is designed to fully exploit partial visual observations to generate lifelike human avatars.
Our proposed diffusion model learns a generative mapping from these partial observations to complete and authentic 3D mesh reconstruction.
Additionally, we introduce a feed-forward mesh refinement network that enhances the fidelity and identity preservation of the generated avatars, eliminating the need for person-specific test-time optimization.
By leveraging a universal prior model that decodes a generated mesh into a set of 3D Gaussians, we generate a photorealistic 3D Gaussian avatar, capable of being driven with novel expressions in real-time.
Our experiments demonstrate that the avatars generated by our feed-forward approach faithfully represent diverse identities and surpass the visual quality of avatars produced by recent competing methods.

\end{abstract}

%% file: sec/1_intro.tex
\section{Introduction}
\label{sec:intro}
Photorealistic human avatars serve as the foundation for enabling 
immersive telepresence in virtual and augmented reality~\cite{lombardi2018deep,ma2021pica,lombardi2021mvp}.
An authentic 3D human avatar that can be represented, recognized, and animated as self can significantly enhance user experience and engagement.
Recent advances in the computer vision and graphics field~\cite{kerbl20233dgs,mildenhall2020nerf,saito2024rgca} have unblocked the creation of highly realistic avatars.
Still, creating such highly realistic and drivable 3D avatars typically requires a cumbersome and time-consuming capture pipeline, which limits the democratization of such promising technologies in reality.

The core challenge of contemporary avatar creation pipelines is the trade-off between the abundance of observation and the computation burden during the capture setup. 
Typically, dense visual observations from accurately calibrated multi-view human performance capture systems~\cite{joo2015panoptic,kirschstein2023nersemble,lombardi2018deep,yu2020humbi,yoon2023humbi,he2024diffrelight} help achieve high-fidelity 
3D/4D avatar reconstruction results while requiring significant computational resources and complex processing pipelines.
On the other hand, using accessible capture methods, \eg, monocular phone capture or profile images, can streamline the capture process but require strong prior knowledge to compensate for the lack of visual evidence.

Recently, a line of work~\cite{chen2022ipica,li2024uravatar} tried
to streamline existing avatar creation pipelines using more casual user inputs, \eg, monocular video captures.
These methods introduced 
the universal prior model (UPM) that covers 
the universal corpus
of human appearances and geometries. 
The UPM gets a canonical 3D mesh representing the target identity, 
and decodes it into a highly detailed, real-time drivable avatar, 
often represented in a set of volumetric primitives~\cite{lombardi2021mvp} or 3D Gaussians~\cite{kerbl20233dgs}. 
While these methods obtained remarkable avatar quality and relaxed the user-side requirements, 
a cumbersome test-time UPM fine-tuning stages are mandatory to balance the evidence-prior trade-offs~\cite{chen2022ipica}.
Moreover, offline 3D head tracking is also required to get reliable conditioning data for the prior models~\cite{li2024uravatar,chen2022ipica}.
Despite the promising quality of the created avatars, these requirements still limit the accessibility of avatar creation to novice users. 

To address these limitations, we propose \textbf{\ours}, a \textbf{F}eed-forward, \textbf{i}nstant Gaussian \textbf{C}odec \textbf{A}vatar creation pipeline.
\ours~takes a casually captured, single portrait image as an input 
and generates an authentic head avatar, represented in a set of 3D Gaussian primitives that can be driven in real-time with arbitrary head pose and expression parameters.

The core of our system is a module-based, feed-forward design that seamlessly connects human-centric vision foundation models, a generative model, and a feed-forward refinement model.
Given a single portrait image, we obtain partial and incomplete visual observations, such as RGB face texture, normal, UV and vertex coordinates, 
by leveraging tailored human-centric vision foundation models~\cite{khirodkar2024sapiens}.
We then perform a diffusion-based generative mapping that converts the partial information into a complete and realistic human avatar, represented as a canonical textured mesh.
With such a cascaded design, we make the most of the visual observation one can get from the pixel space and leverage the generative prior learned from the dataset of high-quality human avatar assets. 
Furthermore, we introduce a feed-forward mesh refinement network to enhance the image-space alignment of the generated avatar.
We found the proposed feed-forward mesh refinement network to be essential in achieving realistic and authentic avatars, 
as it corrects the subtle details such as skin tone and cloth details, which are crucial for the avatar's authenticity.
Finally, the subsequent universal prior model~\cite{chen2022ipica,li2024uravatar} decodes the generated canonical meshes into real-time drivable 3D Gaussian avatars.
Overall, \ours~generates an authentic Codec Avatar from a single image in \emph{5 seconds} in a truly feed-forward manner (\Fref{fig:teaser}).

We evaluate the quality of \ours-generated avatars with unseen, diverse identities and expressions and show that our approach outperforms recent single-image-based avatar generation methods by a large margin visually and quantitatively.
We also investigate the effects of the core design choices.

We summarize our main contributions as follows:
\begin{itemize}
    \item We propose \ours, a feed-forward system for creating authentic human avatars from a single casual portrait image.
    \item We design a diffusion model that generates complete avatar texture and geometry, conditioned on partial observations.
    \item We introduce a feed-forward mesh refinement module, which enhances the fidelity of avatars without involving a person-specific test-time optimization process.
\end{itemize}

%% file: sec/2_related.tex
\section{Related Work}
\label{sec:related}

We aim to build a feed-forward system for creating an authentic facial avatar from a single portrait image with generative modeling. 
We briefly review these lines of work.

\paragraph{Avatar Generation from Monocular Imagery}
Creating life-like 3D facial avatars from monocular images or videos is a highly ill-posed problem. 
Existing methods typically formulate this task as an optimization problem with domain-specific priors to compensate for the missing information from single-view imagery.
Within this paradigm, monocular avatar generation methods can be categorized into video-based and image-based approaches.

The video-based approaches~\cite{chen2022ipica, li2024uravatar,athar2024bridging,giebenhain2024mononphm,grassal2021neural,giebenhain2024npga,xiang2024flashavatar,gafni2021dynamic,zielonka2023insta,bharadwaj2023flare,youwang2026elite} 
leverage effectively multi-view nature~\cite{gao2022dynamic} of the dynamic human face videos to track and obtain coarse 3D face geometry and texture. 
Typically, detailed geometry and texture can be obtained with further optimization of 3D representations, \eg, mesh vertex displacement~\cite{grassal2021neural,bharadwaj2023flare}, neural implicit fields~\cite{gafni2021dynamic,zielonka2023insta}, or 3D Gaussians~\cite{giebenhain2024npga,xiang2024flashavatar}.
While most methods focused on personalized avatar generation, Cao~\etal~\cite{chen2022ipica} introduced the concept of the universal prior model (UPM), a facial texture and geometry prior that covers a universal corpus of identities, frames, and views. 
This universal prior facilitated the universal avatar generation from a monocular video with unprecedented texture and geometry details and has been extended to follow-up works~\cite{athar2024bridging,li2024uravatar}.
Although these approaches demonstrated high-fidelity avatars, they require inevitable offline facial tracking stages, which can take up to a few hours; it necessitates the tracking-free image-based approaches.

\input{figures/tex/system}

The image-based approaches~\cite{gecer2019ganfit,gecer2021fastganfit,lattas2023fitme,hao2024idsculpt,an2023panohead,buehler2024cafca,bai2023ffhquv} aim to generate high-fidelity facial avatars from a more casual input, \eg, a profile image, a portrait image, or even an internet image.
The core benefit of this paradigm is that it can circumvent the need for facial tracking. 
As temporal and multi-view cues are absent, generative priors are employed to compensate for the missing information.
PanoHead~\cite{an2023panohead} trained a tri-plane GAN for unconditional avatar generation and performed GAN inversion optimization~\cite{roich2022pti} for personalized avatar generation. However, it could not generate multi-view consistent and controllable avatars~\cite{lyu2024facelift}.
ID-Sculpt~\cite{hao2024idsculpt} employed the Score-Distillation optimization~\cite{poole2022dreamfusion} to leverage the diffusion model's prior knowledge of human heads, but it is limited in terms of generation speed.

Despite huge progress in both video-/image-based approaches, the prior-based optimization methods suffer from the quality and systematic complexity trade-off. 
In our work, we build a fast feed-forward avatar generation pipeline, 
composed of a diffusion model that directly generates texture and geometry using a single image as a condition.

\paragraph{Feed-forward Avatar Generation}
Learning-based feed-forward avatar generation
is a promising direction for streamlining the complex generation pipeline. 
Early works~\cite{feng2021deca,feng2022trust} introduced the regression-based facial texture and geometry reconstruction, where the results typically showed limited 
texture
and the absence of detailed geometry.
Recent methods~\cite{chu2024gpavatar,tran2023voodoo3d,deng2024portrait4d,deng2024portrait4dv2,chu2024gagavatar,tran2024voodooxp} proposed feed-forward methods to animate a 3D portrait from a single image and driving frames, 
and GPAvatar and GAGAvatar showed promising qualities.
However, GPAvatar exhibits visual artifacts due to the dependency on tri-plane avatar representation and a separate super-resolution module.
GAGAvatar generates avatars with limited expression since they
simulate canonical 3D Gaussian avatars via a learned renderer network.

As a concurrent work, FaceLift~\cite{lyu2024facelift} proposes a two-stage method; first, it generates multi-view images from a single-view image and predicts 3D Gaussian~\cite{kerbl20233dgs} parameters with the transformer-based network. 
While the static reconstruction results may look plausible, they only support dynamic avatars via per-frame 3D estimation from a video. This limits the visual quality with severe temporal jittering, and the avatars cannot be freely controlled by the user.
In contrast, we directly generate a complete Gaussian avatar real-time drivable with any expression signals from the user.

%% file: figures/tex/system.tex
\begin{figure*}[t]
\centering
   \includegraphics[width=\linewidth]{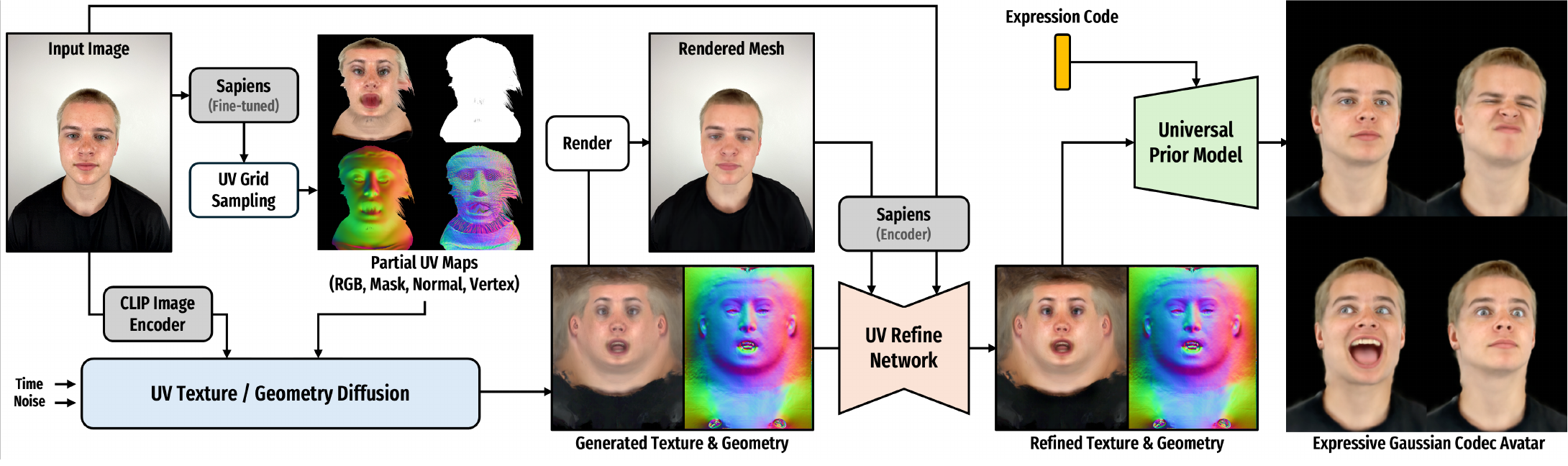}
   \caption{\textbf{\ours: Pipeline Overview.}
   \ours~generates a high-quality drivable Gaussian Codec Avatar from only a single portrait image, without offline face tracking or person-specific fine-tuning. 
   We introduce three main modules: 1) \textcolor{c_diff}{UV texture and geometry diffusion model}, 2) \textcolor{c_urn}{feed-forward UV refinement network}, and 3) \textcolor{c_upm}{universal prior model}.
   \ours~first employs fine-tuned Sapiens~\cite{khirodkar2024sapiens} models to obtain per-pixel UV and vertex coordinates and normal estimation, 
   and unwraps to partial RGB, visibility mask, normal and vertex coordinates in UV space. 
   Then, the \textcolor{c_diff}{diffusion model} takes the partial UV maps, CLIP embedding~\cite{radford2021clip} of the input image, and random noise to generate complete texture and geometry.
   The learned \textcolor{c_urn}{UV refinement network} takes the generated texture and geometry as input, rich visual features of the rendered mesh image and the input image as conditions, and performs feed-forward texture and geometry refinement. 
   Finally, the \textcolor{c_upm}{universal prior model} gets expression codes, mesh texture, and geometry as inputs to generate Gaussian Codec Avatars and drive in real-time.
   }   
\label{fig:pipeline} 
\end{figure*}

%% file: sec/3_method.tex
\section{Feed-forward Gaussian Codec Avatar Generation from a Single Portrait Image}
\label{sec:method}

We introduce \ours, a feed-forward system to generate high-fidelity Gaussian Codec Avatars from a monocular portrait image.
We visualize the \ours~pipeline in \Fref{fig:pipeline}. At a high level, the input is a single portrait image, and the output is a drivable Codec Avatar represented in mesh-aligned 3D Gaussians.
We first introduce
how we leverage vision foundation models and generative modeling to approach this highly ill-posed task in \Sref{sec:diffusion}.
Then, we provide details of the feed-forward refinement module for avatar quality enhancement in \Sref{sec:ff_refinement}.
Finally, we elaborate on the 3DGS decoding and the real-time driving of the avatars in \Sref{sec:upm_driving}.

\subsection{Diffusion-based Avatar Texture and Geometry Generation from a Single Image}
\label{sec:diffusion}

The core module of \ours~is a diffusion model that generates complete mesh texture and geometry of avatars. 
We first introduce the conditioning signals for our diffusion model. 

\paragraph{Foundation Models for Conditioning Diffusion}
\label{sec:foundation_models}
A single portrait image lacks the information for 
complete avatar generation.
Thus, we leverage the prior knowledge of vision foundation models, CLIP~\cite{xu2024metaclip,radford2021clip} and Sapiens~\cite{khirodkar2024sapiens}, to extract rich features and comprehensive information.

Given a portrait image $\bI_\text{ref}$,
we first obtain CLIP image embedding $\textbf{f}_\text{CLIP}$, which encodes visual semantic information of the subject~\cite{radford2021clip}. 
We also use the fine-tuned versions of Sapiens~\cite{khirodkar2024sapiens}, which predict per-pixel UV coordinates of the 3D mesh surface, mesh vertex coordinates, and normals.
Then, pixel RGB values, predicted normal vector, and 
vertex coordinates
are 
unwrapped
into partial UV texture maps using the predicted UV coordinates, resulting in four partial UV maps: $\textbf{UV}_\text{partial}{=}[\textbf{UV}_\text{RGB}, \textbf{UV}_\text{mask}, \textbf{UV}_\text{nrm}, \textbf{UV}_\text{vtx}].$
We use the CLIP embedding $\textbf{f}_\text{CLIP}$ and partial UV maps $\textbf{UV}_\text{partial}$ as the conditions for our diffusion model. 
Please refer to the supplementary Sec.~\colorref{B.1} for 
the Sapiens fine-tuning details.

\paragraph{Mesh as a Proxy Avatar Representation}
While our final avatar representation is 3D Gaussians, we use the generated mesh texture and geometry as a proxy avatar representation.
Inspired by~\cite{chen2022ipica,li2024uravatar}, the generated mesh texture and geometry serve as ID conditioning inputs for generating and driving authentic avatars represented in 3D Gaussians, detailed later in~\Sref{sec:upm_driving}.
Note that prior methods~\cite{chen2022ipica,li2024uravatar} used heuristic offline face tracking to obtain the ID conditioning mesh texture and geometry. In contrast, we directly generate them from just a single image using a diffusion model.

\paragraph{Diffusion-based Texture and Geometry Generation}
We design 
a diffusion model that 
generates complete textures and mesh geometries of avatars from the visual features and partial information. 
Given an image $\bI_\text{ref}$, we obtain
the CLIP image embedding $\textbf{f}_\text{CLIP}$ and partial UV maps $\textbf{UV}_\text{partial}$ (from \Sref{sec:foundation_models}).
We design a latent diffusion model $\mathcal{F}_\theta$ in the DiT architecture~\cite{peebles2023dit,kant2025pippo},
which takes $\textbf{f}_\text{CLIP}$, $\textbf{UV}_\text{partial}$, domain switcher $\bd$~\cite{long2024wonder3d}, 
diffusion timestep $t$ and random noise $\bz$ 
to generate complete UV texture map $\tilde{\bT}\in\mathbb{R}^{H\times W\times 3}$ 
and UV geometry map $\tilde{\bG}\in\mathbb{R}^{H\times W\times 3}$.
We use a pre-trained SDXL VAE~\cite{podell2024sdxl} to encode partial UV maps, texture, and geometry maps into compact latent codes and map them back into the original data space.
For details on the diffusion model's architecture, please refer to the supplementary Sec.~\colorref{B.2}.

We train a single diffusion model $\mathcal{F}_\theta$ for generating both UV texture and geometry map, using the conditional flow matching~\cite{lipman2023flow} objective as follows:
\begin{multline}
    \label{eq:diffusion_objective}
    \mathcal{L}_\text{diffusion}=\lVert\bv_{t}^\text{T} - \mathcal{F}_{\theta}(\bx_{t}^\text{T},\textbf{f}_\text{CLIP},\textbf{UV}_\text{partial},\bd^\text{T},t)\rVert_{2}^{2}\,\,{+}\\
    \lVert\bv_{t}^\text{G} - \mathcal{F}_{\theta}(\bx_{t}^\text{G},\textbf{f}_\text{CLIP},\textbf{UV}_\text{partial},\bd^\text{G},t)\rVert_{2}^{2},
\end{multline}
where the superscripts $\text{T}$ and $\text{G}$ denote the UV texture and geometry domains, 
$\bv_\text{t}^\text{*}$ denotes the ground-truth flow field, derived by the optimal transport formulation of conditional flow matching~\cite{lipman2023flow}, 
$\bx_\text{t}^\text{*}$ denotes the noise-added latents at diffusion timestep $t$ for texture and geometry maps, 
and $\bd^\text{*}$ is the constant, domain switcher~\cite{long2024wonder3d} that decides which UV domain (texture or geometry) to denoise for.

Note that the target task of our diffusion model is different from that of diffusion-based inpainting models~\cite{rombach2022high,lugmayr2022repaint}.
For casual input images, the partial UV maps obtained via Sapiens models and UV grid sampling are typically noisy as they cannot infer the accurate texture and geometries for self-occluded regions, 
\eg, mouth interior, space between chin and neck, or subject's boundaries. 
Moreover, UV and vertex coordinate
prediction from a single image is a highly ill-posed problem with a risk of imperfection. 
Thus, we cannot simply trust the predictions and inpaint only for the missing parts.
Instead, our diffusion model is trained to imagine complete texture and geometries from imperfect observations while preserving the ID information from the image. 
We embody this ability by training the diffusion model with large-scale texture and geometry assets of 3D humans, accurately collected from phone captures and high-end multi-view capture domes~\cite{chen2022ipica,saito2024rgca,lombardi2021mvp,li2024uravatar,athar2024bridging}.

\input{figures/tex/refine_example}

\subsection{Feed-forward UV Refinement Network}
\label{sec:ff_refinement}

While the generated avatar in a textured mesh may already look plausible, we further enhance the fidelity and identity (ID) preservation of the avatar.
We observe image-level misalignment between the reference image and the rendered avatars in meshes in the pixel space (see \Fref{fig:refine_ex}).
To generate an authentic avatar, we introduce a subsequent, learned network for texture and geometry refinement, which operates in a feed-forward manner.
Note that we do not perform person-specific test-time optimization~\cite{chen2022ipica,li2024uravatar,buehler2024cafca,hao2024idsculpt} to align the avatar with the image and to enhance the quality.

\paragraph{Learned UV Refinement using Sapiens Features}
We propose a UV refinement network $\mathcal{R}_\phi$ that gets initial texture $\tilde{\bT}$ and geometry map $\tilde{\bG}$ generated from the diffusion model,
and refines them by leveraging the rich visual features of the input image and the rendered mesh image.

Given an input image $\bI_\text{ref}$ (\Fref{fig:refine_ex}\colorref{a}) and the rendered image of the diffusion-generated avatar $\tilde{\bI}_\text{rdr}$ (\Fref{fig:refine_ex}\colorref{b}), 
we extract dense visual features $\textbf{f}_\text{ref}$ and $\textbf{f}_\text{rdr}$ from the Sapiens ViT encoder~\cite{khirodkar2024sapiens}.
As Sapiens ViT encoder is pre-trained with masked-autoencoding task~\cite{he2022maskedautoencoder} with million-scale human-centric images, 
we expect the features $\textbf{f}_\text{ref}$ and $\textbf{f}_\text{rdr}$ to provide informative cues for minimizing the photometric error between the images.
We design the refinement network $\mathcal{R}_\phi$ in U-Net architecture with cross-attention layers~\cite{patrick2022diffusers}.
We choose the cross-attention layer, as the goal of the refinement network is to refine UV maps by referencing the conditions from the different modality, \ie, image features.

Given initial UV texture and geometry maps as inputs, $\tilde{\bT}$ and $\tilde{\bG}$, the refinement network performs cross-attention between UV space and Sapiens features to produce the final texture and geometry maps $\bT$ and $\bG$ as:
$[\bT,\bG]{=}\mathcal{R}_\phi([\tilde{\bT},\tilde{\bG}];\textbf{f}_\text{ref},\textbf{f}_\text{rdr})$.
After obtaining the final UV maps, we 
get the final avatar that is better aligned to the input image (see \Fref{fig:refine_ex}\colorref{d}).
During training, we use the ground-truth face pose parameter and expression code to overlay the mesh on the image and compute the image space loss. At inference time, we may use an off-the-shelf 
regressor, \eg \cite{danecek2022emoca,retsinas2024smirk}, to estimate the parameters.

\paragraph{Training UV Refinement Network} 
We train $\mathcal{R}_\phi$ using triplets of $\{\bI_\text{ref}, \tilde{\bI}_\text{rdr}, [\tilde{\bT},\tilde{\bG}]\}$.
For the training objective, we use a weighted sum of L1 photometric loss, 2D keypoint loss, and mask loss on image space and regularization losses for UV texture and geometry maps as follows:
\begin{equation}
\label{eq:refinenet_loss}
    \mathcal{L}_\text{refine}{=}{\lambda_\text{pho}}\mathcal{L}_\text{pho}{+}\lambda_\text{mask}\mathcal{L}_\text{mask}
    {+}\lambda_\text{kpts}\mathcal{L}_\text{kpts}{+}\lambda_\text{reg}\mathcal{L}_\text{reg},
\end{equation}
where $\mathcal{L}_\text{pho}=\lVert\bI_\text{ref}-\bI_\text{rdr}\rVert_{1}$, $\mathcal{L}_\text{mask}=\lVert\bm_\text{ref}-\bm_\text{rdr}\rVert_{1}$, $\mathcal{L}_\text{kpts}=\lVert\bk_\text{ref}-\bk_\text{rdr}\rVert_{1}$, $\mathcal{L}_\text{reg}=\lVert\tilde{\bT}-\bT\rVert_{2}+\lVert\tilde{\bL}-\bL\rVert_{2}+\lVert\tilde{\bN}-\bN\rVert_{2}$, respectively.
Here, $\bm_\text{ref}$ is the human segmentation mask of $\bI_\text{ref}$, obtained by an off-the-shelf matting model~\cite{shanchuan2022rvm}, 
$\bm_\text{rdr}$ is the mesh foreground mask rendered by a differentiable rasterizer~\cite{pidhorskyi2024rasterized}, 
$\bk_\text{ref}$ is the ground-truth 2D keypoints, and $\bk_\text{rdr}$ is the 2D projected positions of the keypoint-corresponding mesh vertices.
Also, $\tilde{\bL}$, $\bL$, $\tilde{\bN}$ and $\bN$ denote the Laplacian matrix and normal maps of the meshes $\tilde{\bG}$, $\bG$, respectively.
The regularization terms encourage the refined UV maps to not deviate too much from the initial UV maps, preventing the network from overfitting only for the visible parts.

\subsection{Decoding Mesh into Drivable Gaussian Codec Avatar via Universal Prior Model}
\label{sec:upm_driving}
Given the generated textured mesh as a proxy representation for our avatar, we convert the mesh into a set of 3D Gaussians as a final representation.
We choose 3D Gaussians due to its efficiency and expressiveness in representing details.
Inspired by prior work~\cite{chen2022ipica,li2024uravatar}, we use a hypernetwork-based 3D Gaussian avatar generation model called the Universal Prior Model (UPM). 
We use the UPM to decode the generated meshes into high-fidelity drivable 3D Gaussian avatars.

The UPM consists of two modules, $\mathcal{U}_\psi=\{\mathcal{E}_{\psi_\text{id}},\mathcal{D}_{\psi_\text{dec}}\}$, where 
$\mathcal{E}_{\psi_\text{id}}$ refers to the identity encoder, and $\mathcal{D}_{\psi_\text{dec}}$ refers to the 3D Gaussian decoder.
For training and dataset details, please refer to the supplementary Sec.~\colorref{B.4},
and
\cite{chen2022ipica,li2024uravatar}.
The identity encoder $\mathcal{E}_{\psi_\text{id}}$ is a CNN-based hypernetwork~\cite{ha2017hypernetworks} that takes ID conditioning mesh
in the form of UV texture and geometry maps, $\bT$ and $\bG$,
and generates ID-specific bias maps, $\Psi_\text{id}$, for the 3D Gaussian decoder.
We obtain $\bT$ and $\bG$ from the previous diffusion generation and feed-forward refinement stage.
The generated bias maps $\Psi_\text{id}$ serve as the modulation signal for the decoder layers. 
The decoder $\mathcal{D}_{\psi_\text{dec}}$ takes the bias maps,
along with the driving signals to produce a set of 3D Gaussians that represent a Codec Avatar.
For driving signals, we use expression codes $\mathbf{e}$, view- and gaze-direction vectors, $\mathbf{v}$ and $\mathbf{g}$, following~\cite{chen2022ipica}.

In summary, given UV texture and geometry maps, $\bT$ and $\bG$, generated from the diffusion model and refinement network,
we generate Codec Avatars in any expression, represented with a set of 3D Gaussians as follows:
\begin{align*}
    \Psi_\text{id} &= \mathcal{E}_{\psi_\text{id}}(\bT, \bG),\\
    \{\delta\bx,\delta\bc,\bq,\bs,\bo\}&=\mathcal{D}_{\psi_\text{dec}}(\mathbf{e},\mathbf{v},\mathbf{g},\Psi_\text{id}),
\end{align*}
where $\delta\bx$ and $\delta\bc$ denote the position and color offsets of 3D Gaussians from the mesh surface position and color, and $\bq$, $\bs$, and $\bo$ denote the rotation, scale and opacity parameters for each 3D Gaussian primitive~\cite{kerbl20233dgs}.

%% file: figures/tex/refine_example.tex
\begin{figure}[t]
\centering
\includegraphics[width=\linewidth]{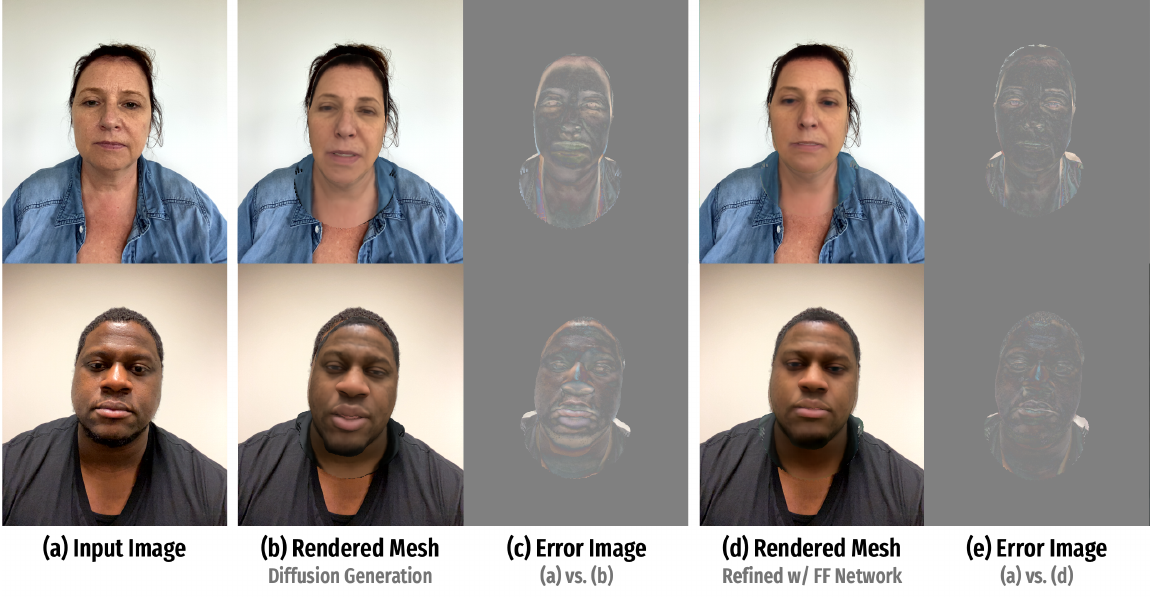}
\caption{\textbf{Effect of Feed-forward UV Refinement.} Our UV refinement network uses rich image features from (a) input image and (b) rendering of diffusion generated mesh
to refine the mesh texture and geometry, resulting in enhanced avatar fidelity and ID preservation (d). For error images, the gray area means zero error.
}\vspace{5mm}
\label{fig:refine_ex}
\end{figure}

%% file: sec/4_exp.tex
\section{Experiments}
\label{sec:exp}

We first introduce the train and test datasets.
Then, we provide visualizations of our generated avatars 
and compare \ours~with the recent competing methods.
We also conduct ablation studies to support our core design choices.

\input{figures/tex/data_sample}

\input{figures/tex/qual}

\subsection{Datasets}
\label{sec:dataset}
To train our diffusion model, we need pairs of \{portrait image, UV texture/geometry map\} (see inset).
Note that we show the geometry UV map in the style of a normal map, just for visualization.
We obtain these data from
two 
heterogeneous datasets: 1) multi-view dome captured dataset and 2) iPhone captured dataset.
Following Cao~\etal~\cite{chen2022ipica}, we use a multi-view dome to capture dynamic facial performance, track meshes, and unwrap texture and geometry UV maps. We obtain portrait images by choosing frames from a face-looking camera.
For iPhone captures, we obtain portrait images from rear-camera frames, track meshes, and unwrap UV texture/geometry maps from monocular videos.
We collect total 1,948 identities (IDs) for the dome dataset and split into 1,932 train and 16 test IDs. For the iPhone dataset, we collect 12,539 IDs, split into 12,439 train and 100 test IDs.

For training the feed-forward UV refinement network, we render the diffusion generation results and build pairs of \{portrait image, generated mesh image\}, and train the model in a self-supervised manner (\Eref{eq:refinenet_loss}).
\vspace{-1.5mm}

\subsection{Qualitative Results}
In \Fref{fig:qual}, we visualize the generated 3D Gaussian avatars for unseen test IDs.
To show \ours's generalization capability, we choose test IDs with diverse human attributes, including races, genders, ages, and hairstyles. 
Given only a single portrait image and random driving expressions, our method generates realistic and ID-preserving Gaussian avatars.
From the results, we observe that the visual details in the input image, \eg, tattoos or necklace, are well reflected in the generated 3D Gaussian avatars.
As we use the vision foundation models to obtain the conditioning data for the diffusion model,
\ours~pipeline is robust to the 
input image characteristics, such as the body coverage in the image and the camera's position.
Furthermore, \ours~can \emph{imagine} the unobserved facial areas, \eg, interior mouth or eye pupil, and reasonably generates the missing textures and geometries, thanks to our conditional diffusion formulation.

\input{figures/tex/static_comparison}

\input{figures/tex/reenact}

\subsection{Comparison with Competing Methods}

\paragraph{Competing Methods}
We compare \ours's textured mesh generation quality with PanoHead.
PanoHead reconstructs full 3D head avatar from a single portrait, via 3D-aware GAN inversion optimization~\cite{roich2022pti} (takes $\sim$80 secs per image).

We compare the quality of \ours's avatars under dynamic expressions with 
recent monocular 3D avatar animation methods: GPAvatar~\cite{chu2024gpavatar}, VOODOO~3D~\cite{tran2023voodoo3d}, Portrait4D-v1/v2~\cite{deng2024portrait4d,deng2024portrait4dv2} and GAGAvatar~\cite{chu2024gagavatar}.
Given a source image, each method generates an avatar in various 3D representations, \eg, tri-plane or 3D Gaussians.
The avatare are animated using offline tracked FLAME~\cite{li2017flame} meshes.
For all the comparisons, we use held-out test IDs from our datasets.

\paragraph{Static Avatar Reconstruction Comparison}
In \Fref{fig:static}, we compare the visual quality of the generated static head avatars from PanoHead~\cite{an2023panohead} and our method.
PanoHead and \ours~take only a single portrait image as an input and generate avatars in tri-plane and textured mesh, respectively. 
\ours~generates more realistic and view-consistent complete head avatars than PanoHead. 
Specifically, PanoHead suffers from severe visual artifacts such as ghost face or floaters for side or back views,
whereas our method can create view-consistent and realistic face texture and geometry.
Also, PanoHead requires per-image GAN inversion to obtain personalized latent codes, which takes about $15\times$ longer execution time than our feed-forward generation pipeline.
More importantly, avatars generated via PanoHead remain static and cannot be animated as they are not anchored with controllable expression parameters. 
In contrast, our mesh-based avatars can be animated in real-time with arbitrary expression codes obtained from tracking~\cite{chu2024gagavatar,qian2024gaussianavatars,youwang2024neuface}, head-mounted cameras~\cite{bai2024universal}, or multi-modal generative models~\cite{ng2024audio2photoreal}.

\paragraph{Animated Avatar Comparison}
We evaluate the animation quality of the generated avatars and compare with the recent competing methods~\cite{chu2024gpavatar,tran2023voodoo3d,deng2024portrait4d,deng2024portrait4dv2,chu2024gagavatar}.
For \ours, we first generate canonical Gaussian avatars for the unseen test ID using our feed-forward pipeline. 
We use 16 held-out IDs from our dome capture dataset, covering diverse races, genders, and hairstyles.
We tracked per-frame expression codes for each test ID with corresponding video frames (total $\sim$1,500 frames).
Finally, we drive the generated avatar using the unseen expression codes, \ie, zero-shot animation.
For competing methods,
we follow their 3D face tracking protocol and pipeline to animate their generated avatars for test IDs using the driving video sequence (see \Fref{fig:reenact}).

\begin{table}[t]
    \caption{\textbf{Quantitative Comparison: Animated Avatar.} 
    We evaluate the animation quality of the generated avatars using recent competing methods and \ours~(mesh \& 3DGS). For pairs of input portrait images and facial videos of 16 held-out IDs, avatars created by \ours~show superior photometric quality and ID preservation.
    }\vspace{-1.5mm}
    \centering
    \resizebox{\linewidth}{!}{
    \begin{tabular}{lcccc}
    \toprule
         & PSNR ($\uparrow$) & SSIM ($\uparrow$) & LPIPS ($\downarrow$) & ID-CSIM ($\uparrow$)\\
     \cmidrule{1-5}
         GPAvatar~\cite{chu2024gpavatar} & 
         19.565 & 
         \cellcolor{tabthird}0.7648 & 
         0.1915 & 
         0.3166 
         \\
         VOODOO~3D~\cite{tran2023voodoo3d} &
         19.321 & 
         0.6983 & 
         0.2756 & 
         0.4339 
         \\
         Portrait4D-v1~\cite{deng2024portrait4d} &
         15.006 & 
         0.3743 & 
         0.4138 & 
         0.2135
         \\
         Portrait4D-v2~\cite{deng2024portrait4dv2} &
         15.704 & 
         0.3871 & 
         0.3765 & 
         0.2545
         \\
         GAGAvatar~\cite{chu2024gagavatar} & 
         \cellcolor{tabthird}22.157 & 
         0.7513 & 
         \cellcolor{tabfirst}0.1320 &
         \cellcolor{tabthird}0.3522
         \\
         \cmidrule{1-5}
         \textbf{\ours\textsubscript{\emph{Ours/Mesh}}} & 
         \cellcolor{tabsecond}24.281 & 
         \cellcolor{tabsecond}0.9625 & 
         \cellcolor{tabthird}0.1381 & 
         \cellcolor{tabsecond}0.5233 
         \\
         \textbf{\ours\textsubscript{\emph{Ours/3DGS}}} & 
         \cellcolor{tabfirst}24.508 & 
         \cellcolor{tabfirst}0.9637 & 
         \cellcolor{tabsecond}0.1365 & 
         \cellcolor{tabfirst}0.5867 
         \\
    \bottomrule
    \end{tabular}
    }\vspace{-3mm}
    \label{tab:mugsy_quant}
\end{table}

In \Tref{tab:mugsy_quant}, we report the photometric reconstruction metrics, \ie, PSNR, SSIM, and LPIPS.
We compute these metrics between the ground-truth face capture frames and the renderings of the animated generated avatars by each method.
We compute metrics only for the face region to avoid influence from the background.
\ours~outperforms all the competing methods
in terms of PSNR and SSIM and shows a comparable score with GAGAvatar in LPIPS.
We also evaluate 
the ID preservation and report the ID similarity metric (ID-CSIM). 
We compute the cosine similarity of ArcFace~\cite{deng2019arcface} embedding extracted from the source portrait image and the renderings of generated dynamic avatars. 
We use DeepFace implementation for computing ID-CSIM~\cite{serengil2021lightface}.
\ours~achieves a higher ID-CSIM score than the other methods, 
supporting the superiority of our method in generating ID-preserving, authentic avatars from a single image.

\input{figures/tex/ablation}

\begin{table}[t]
    \caption{\textbf{Ablation Study: Quantitative.} We evaluate the quality of generated avatars by ablating core design components. 
    Our diffusion model conditioned with partial observations from Sapiens~\cite{khirodkar2024sapiens}, semantic features from CLIP~\cite{radford2021clip}, and the feed-forward refinement network helps achieve the highest quality avatars.
    }
    \centering
    \resizebox{\linewidth}{!}{
    \begin{tabular}{ccc c ccc}
        \toprule
            \multicolumn{3}{c}{\textbf{Diffusion Model Config.}} & \multicolumn{1}{c}{\textbf{FF Ref. Net.}} & \multicolumn{3}{c}{\textbf{Metrics}} \\
        \cmidrule(lr){1-3} \cmidrule(lr){4-4} \cmidrule(lr){5-7}
             $\textbf{UV}_\text{RGB}$ & $\textbf{UV}_\text{nrm,vtx}$ & $\mathbf{f}_\text{CLIP}$ & - & PSNR ($\uparrow$) & SSIM ($\uparrow$) & LPIPS ($\downarrow$)\\
         \cmidrule{1-7}
         \cmidrule(lr){1-2} \cmidrule(lr){3-4} \cmidrule(lr){5-7}
             \checkmark &      -     &      -     &      -     & 19.504 & 0.8140 & 0.1806 \\
             \checkmark & \checkmark &      -     &      -     & \cellcolor{tabthird}19.644 & \cellcolor{tabthird}0.8164 & \cellcolor{tabthird}0.1667 \\
             \checkmark & \checkmark & \checkmark &      -     & \cellcolor{tabsecond}19.738 & \cellcolor{tabsecond}0.8431 & \cellcolor{tabsecond}0.1648 \\
             \checkmark & \checkmark & \checkmark & \checkmark & \cellcolor{tabfirst}22.282 & \cellcolor{tabfirst}0.8804 & \cellcolor{tabfirst}0.1569 \\
        \bottomrule
    \end{tabular}
    }
    \label{tab:ablation}
\end{table}

The qualitative comparison results in \Fref{fig:reenact} show that our method generates more authentic avatars, especially for skin tones and extreme facial expressions, even for zero-shot test IDs.
GPAvatar produces severe visual artifacts on the avatars, possibly caused by their implicit avatar representation of tri-plane+MLPs and subsequent super-resolution module.
GAGAvatar shows better quality than GPAvatar but still suffers from the limited expressivity of the generated avatars. 
We postulate this is because GAGAvatar does not directly infer dynamic avatars in explicit 3D Gaussians. GAGAvatar estimates a canonical 3D Gaussian avatar and uses its learned neural renderer to simulate dynamic avatars, which may not generalize well to extreme expressions.

\subsection{Ablation Study}

In \Fref{fig:ablation}, we visualize the effects of our core system design choices.
We show 3D Gaussian Avatars in the neutral pose and expression and in textured 3D meshes reposed to match the subject in the image.
We mainly investigate the effects of conditioning information for the diffusion model.
In \Fref{fig:ablation}\colorref{a}, we show the avatar generated with the diffusion model, trained to generate complete texture and geometry only from a partial UV RGB texture map. 
As pixel values of RGB UV maps are insufficient to reason about the geometries of a subject, severe identity shift and geometry misalignment occur.
By adding geometry cues with normal and 3D vertex UV maps as conditions, we obtain an improved avatar with reasonable geometry (\Fref{fig:ablation}\colorref{b}). 
Then, with the CLIP embedding injected as a conditioning signal for the diffusion model, we obtain details such as hood (\Fref{fig:ablation}\colorref{c}).
Finally, by adding the subsequent feed-forward UV refinement network (\Sref{sec:ff_refinement}), 
we obtain a high-quality avatar with the vivid skin tone and realism (\Fref{fig:ablation}\colorref{d}).
In \Tref{tab:ablation}, we quantitatively compare the model design variants on 100 iPhone capture held-out test IDs,
and the metrics align with the visual differences.

%% file: figures/tex/data_sample.tex
\begin{wrapfigure}[10]{R}{0.5\linewidth}
\centering
    \vspace{-4mm}
    \hspace{-4mm}\includegraphics[width=1\linewidth]{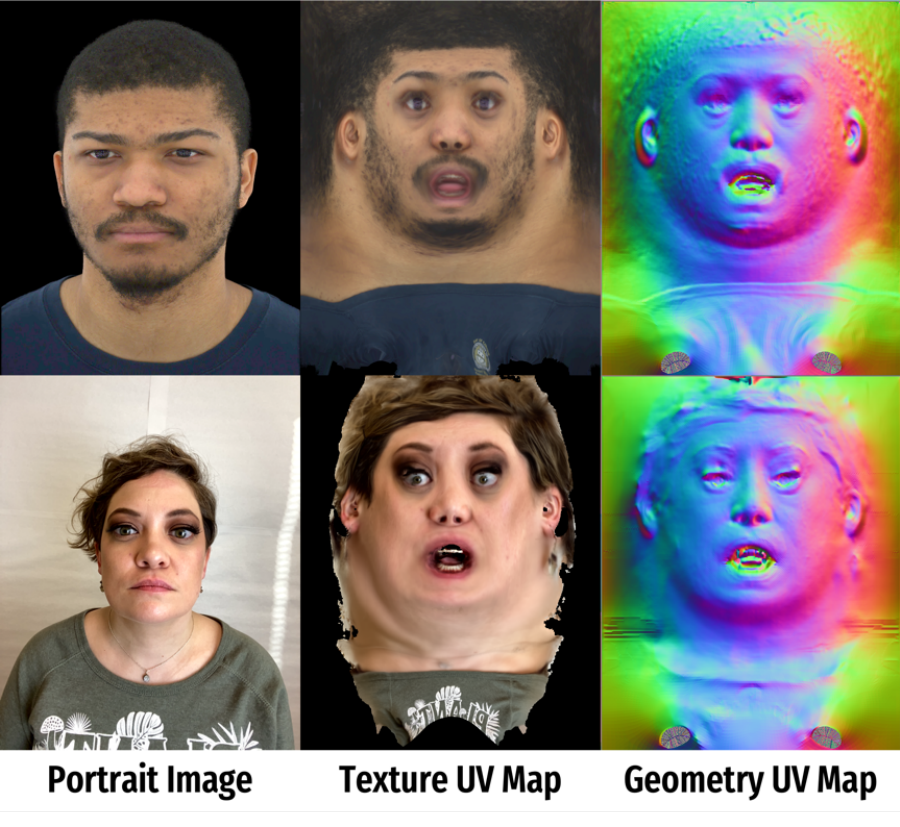}\hspace{-2mm}\vspace{-5mm}
    \label{fig:datasample}
\end{wrapfigure}

%% file: figures/tex/qual.tex
\begin{figure*}[ht]
    \centering
       \includegraphics[width=1\linewidth]{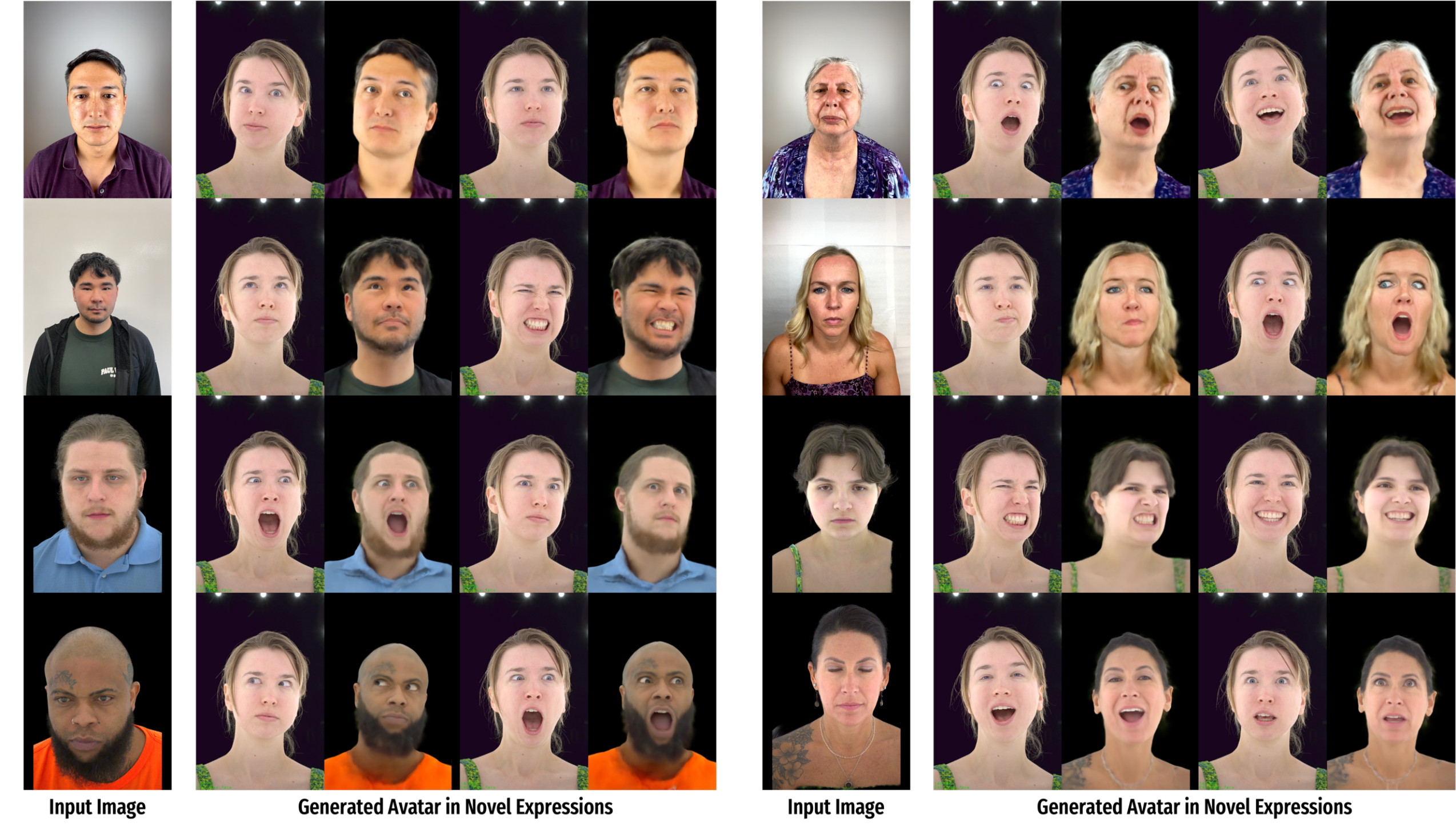}\vspace{-1.5mm}
       \caption{\textbf{Qualitative Results}.
       We show the animated results of our generated 3D Gaussian avatars for test IDs and novel expressions.
       Our \ours~generates authentic, ID-preserving avatars for diverse attributes, \eg, races, genders, ages, hairstyles, and expressions, only from a single image.
       Also, the input image's visual details, such as tattoos or accessories, are faithfully reflected in the 3D Gaussian avatars.
       Note that \ours~can generate unseen observations from the input image, such as the mouth interior and eye pupil, aided by our diffusion model.
       Please refer to the supplementary video for the dynamic avatar animation results.
       }\vspace{-2mm}
    \label{fig:qual} 
\end{figure*}

%% file: figures/tex/static_comparison.tex
\begin{figure}[t]
\centering
\includegraphics[width=0.95\linewidth]{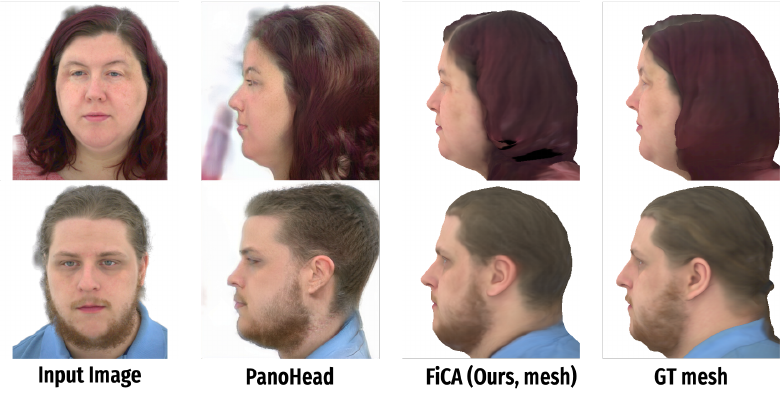}
\vspace{-3mm}
\caption{\textbf{Qualitative Comparison: Static Avatar.} 
PanoHead takes $\sim$80 secs. to generate an avatar with per-subject GAN inversion.
For \ours~(ours), we visualize the textured meshes, which takes $\sim$5 seconds to generate. 
\ours~shows better completeness, especially for extreme viewpoints.
Note that the \ours~meshes are later decoded into animatable 3D Gaussians with visual details.
}
\label{fig:static}
\end{figure}

%% file: figures/tex/reenact.tex
\begin{figure*}[t]
\centering
\includegraphics[width=\linewidth]{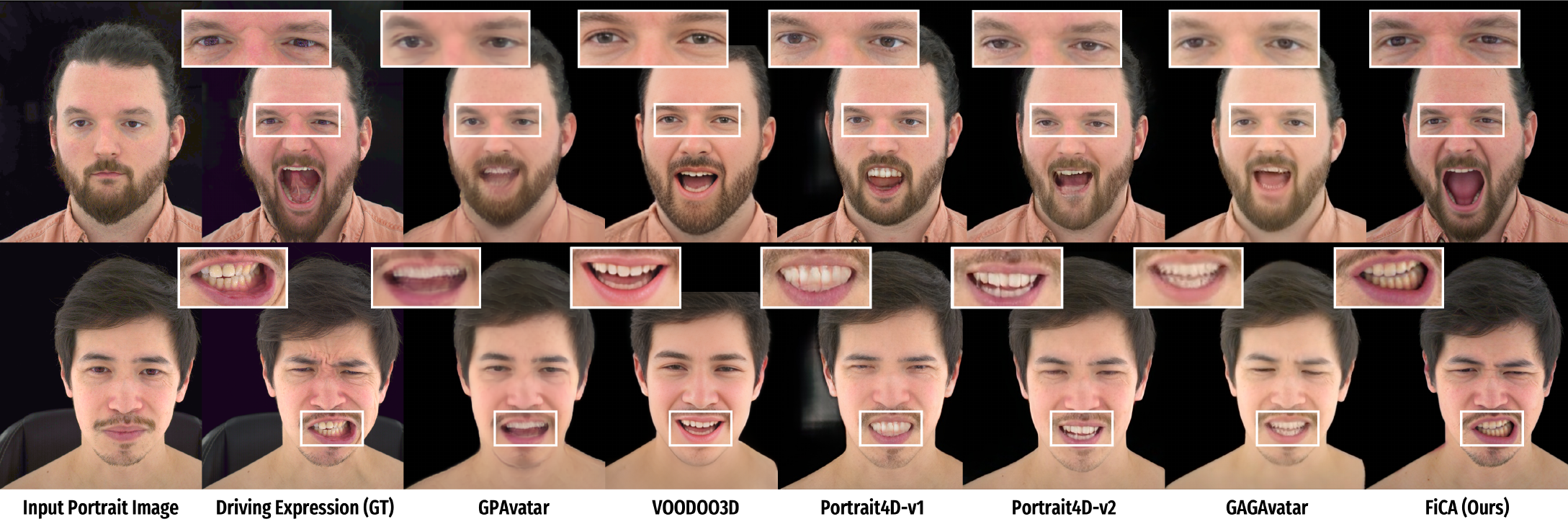}\vspace{-3mm}
\caption{\textbf{Qualitative Comparison: Animated Avatar.} We compare \ours~with recent 3D portrait animation methods~\cite{chu2024gpavatar,tran2023voodoo3d,deng2024portrait4d,deng2024portrait4dv2,chu2024gagavatar}. 
Given an input portrait image of held-out identity, we generate avatars using all methods and drive them using 
tracked driving expression codes of the same identity.
\ours~shows better avatar rendering quality, especially for extreme expressions and skin tones.
}\vspace{-1.5mm}
\label{fig:reenact}
\end{figure*}

%% file: figures/tex/ablation.tex
\begin{figure}[t]
\centering
\includegraphics[width=\linewidth]{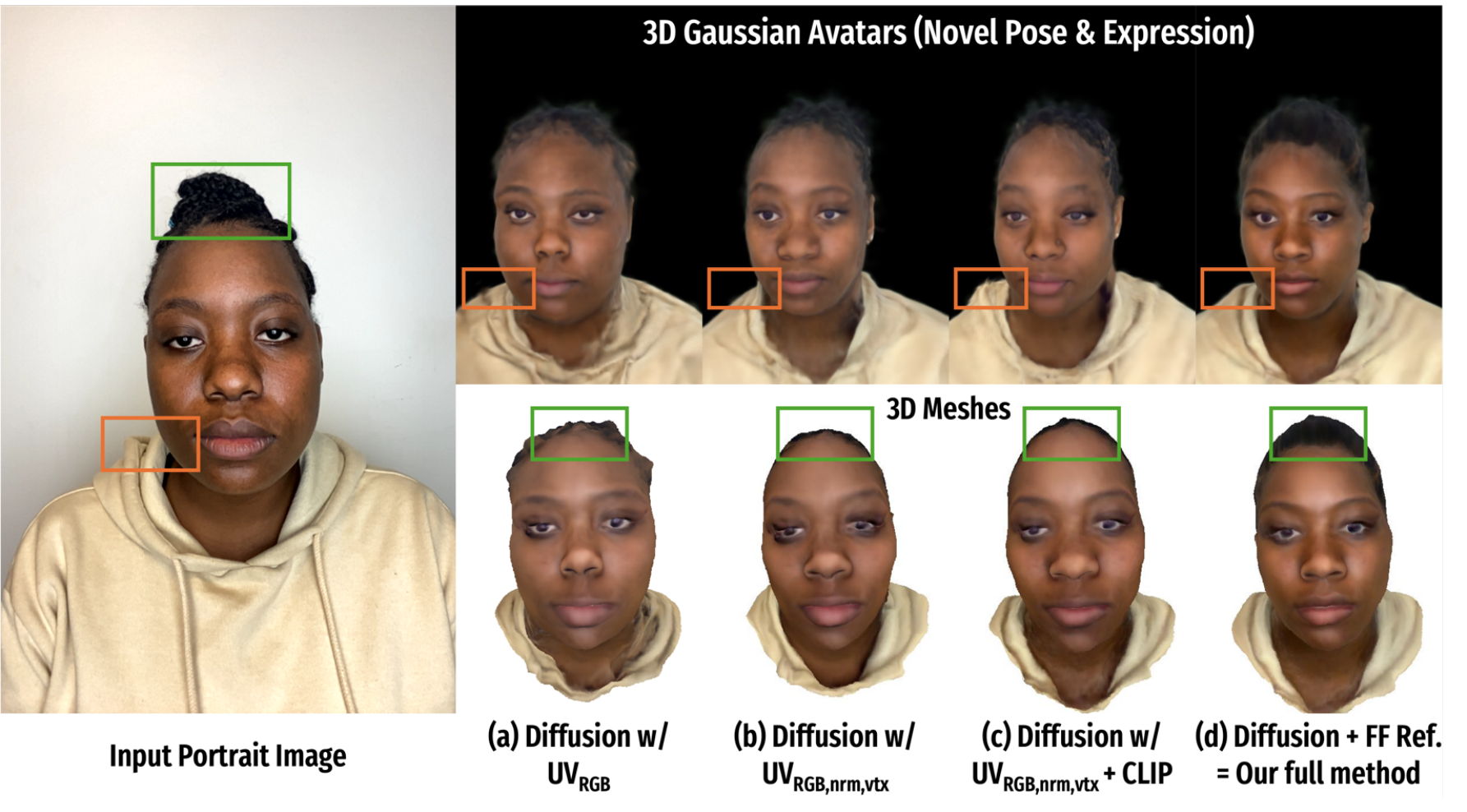}\vspace{-1.5mm}
\caption{\textbf{Ablation Study: Qualitative.}
We visualize the effects of the design choices of \ours.
Compared to the (a) diffusion model trained with only a
partial RGB texture map, 
the (b) diffusion model trained with partial UV maps of normal and 3D vertex
estimation helps achieve the person-specific geometric details,
while (c) adding CLIP image embedding improves the details, such as the hood.
Adding our feed-forward UV refinement network (d) helps achieve the best quality avatar with realistic skin tone and geometries.
}
\label{fig:ablation}
\end{figure}

%% file: sec/5_conc.tex
\input{figures/tex/stylized_avatar}

\section{Conclusion, Discussion and Limitations}
\label{sec:conclusion}

We present \ours, a feed-forward system to generate an authentic Gaussian Codec Avatar from a single image. 
Our system 
connects human-centric vision foundation models with a diffusion model to 
generate complete head texture and geometry.
Our feed-forward texture/geometry refinement network further improves the fidelity and ID preservation of generated avatars.
\ours~shows remarkable avatar generation and animation quality 
for diverse IDs and
novel expressions.

As a promising use-case, we can consider feed-forward editing of Gaussian Codec Avatars. 
Given a portrait image, we can use a 2D image editing method, 
\eg, \cite{brooks2023instructpix2pix,zhang2023magicbrush}, 
to generate stylized portrait image, and use \ours~to generate drivable Codec Avatar (see \Fref{fig:stylize}). 
This can enable efficient 3D avatar stylization and editing paradigms without the need for heuristic 
optimization or manipulation in the 3D domain.

Currently, \ours~can be vulnerable to extreme visual artifacts that may present in input portrait images, \eg, extreme lighting or motion blur.
For our diffusion model, embodying a learned light normalization capability or blur correction for texture maps could be an interesting research problem.
Moreover, extending \ours~to support the joint generation of layered texture and geometry for accessories, \eg, glasses~\cite{li2023megane}, from a portrait image
would be a promising future direction.

%% file: figures/tex/stylized_avatar.tex
\begin{figure}[t]
\centering
\includegraphics[width=\linewidth]{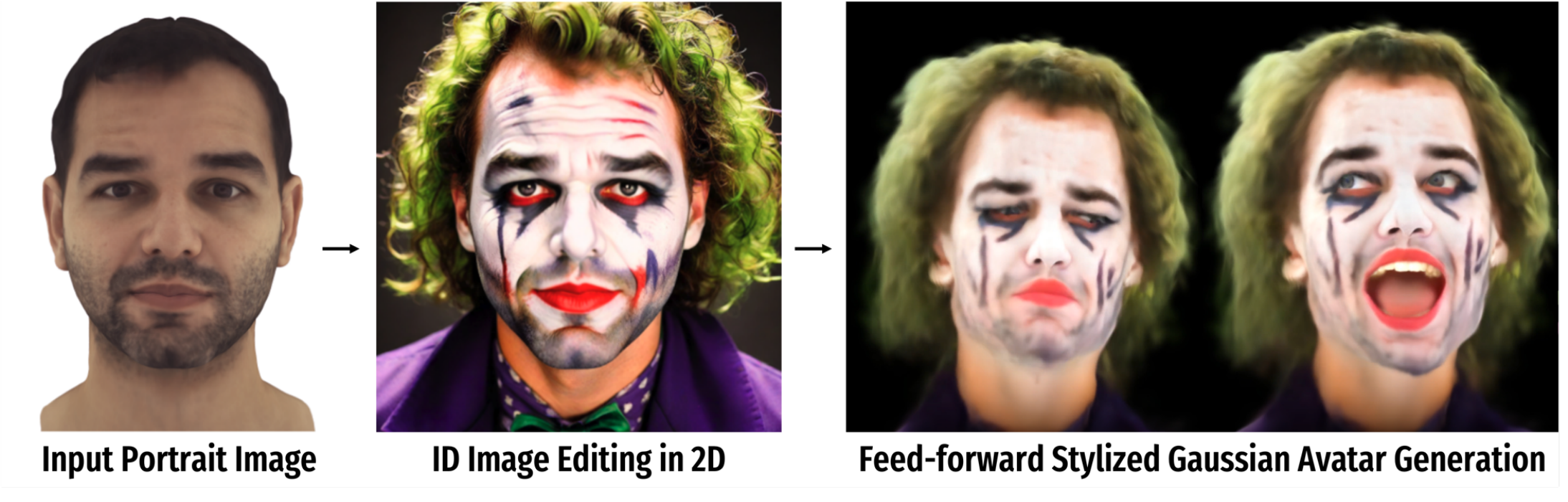}\vspace{-3mm}
\caption{\textbf{Application: Feed-forward Avatar Editing.}
We showcase an application scenario of \ours.
Given an input portrait image, we can use a 2D image editing method to edit images in 2D.
Our feed-forward pipeline creates stylized and drivable Gaussian avatars without
heuristic 3D space optimization or manipulation.
}\vspace{-1.5mm}
\label{fig:stylize}
\end{figure}

%% file: supp.tex
\maketitlesupplementary

\newtheorem{assume}{Assumption}
\newtheorem{definition}{Definition}
\newtheorem{lemma}{Lemma}

\setcounter{section}{0}
\setcounter{figure}{0}
\setcounter{table}{0}
\setcounter{equation}{0}

\renewcommand\thesection{\Alph{section}}
\renewcommand\thefigure{S\arabic{figure}}
\renewcommand{\thetable}{S\arabic{table}}
\renewcommand\theequation{\alph{equation}}
\newcommand{\cmark}{\ding{51}}%
\newcommand{\xmark}{\ding{55}}%
\newcommand{\omark}{\ding{110}}%

In this supplementary material, we provide additional details and results for \ours~that are not
included in the main paper due to the space limit.
Also, we \underline{\textbf{encourage readers to watch the attached video}}, where we show dynamic avatar visualizations.

\makeatletter
\renewcommand{\numberline}[1]{#1\hspace{1.0em}}
\makeatother

\vspace{3mm}
\begingroup
\hypersetup{linkcolor=black}

\noindent\rule{\linewidth}{1.0pt}   %
\vspace{-2mm}
\tableofcontents
\vspace{4mm}
\noindent\rule{\linewidth}{1.0pt}   %

\endgroup

\section{Video for Summary \& Visual Results}
In the attached video, we provide the following content:
\begin{itemize}
    \item \ours~overview and how it works.
    \item Videos of avatars generated by \ours.
    \item Visual comparisons w/ competing methods~\cite{chu2024gagavatar,deng2024portrait4d,deng2024portrait4dv2,tran2023voodoo3d}.
\end{itemize}

\section{More Results}

\begin{figure}[t]
    \centering
    \includegraphics[width=1.0\linewidth]{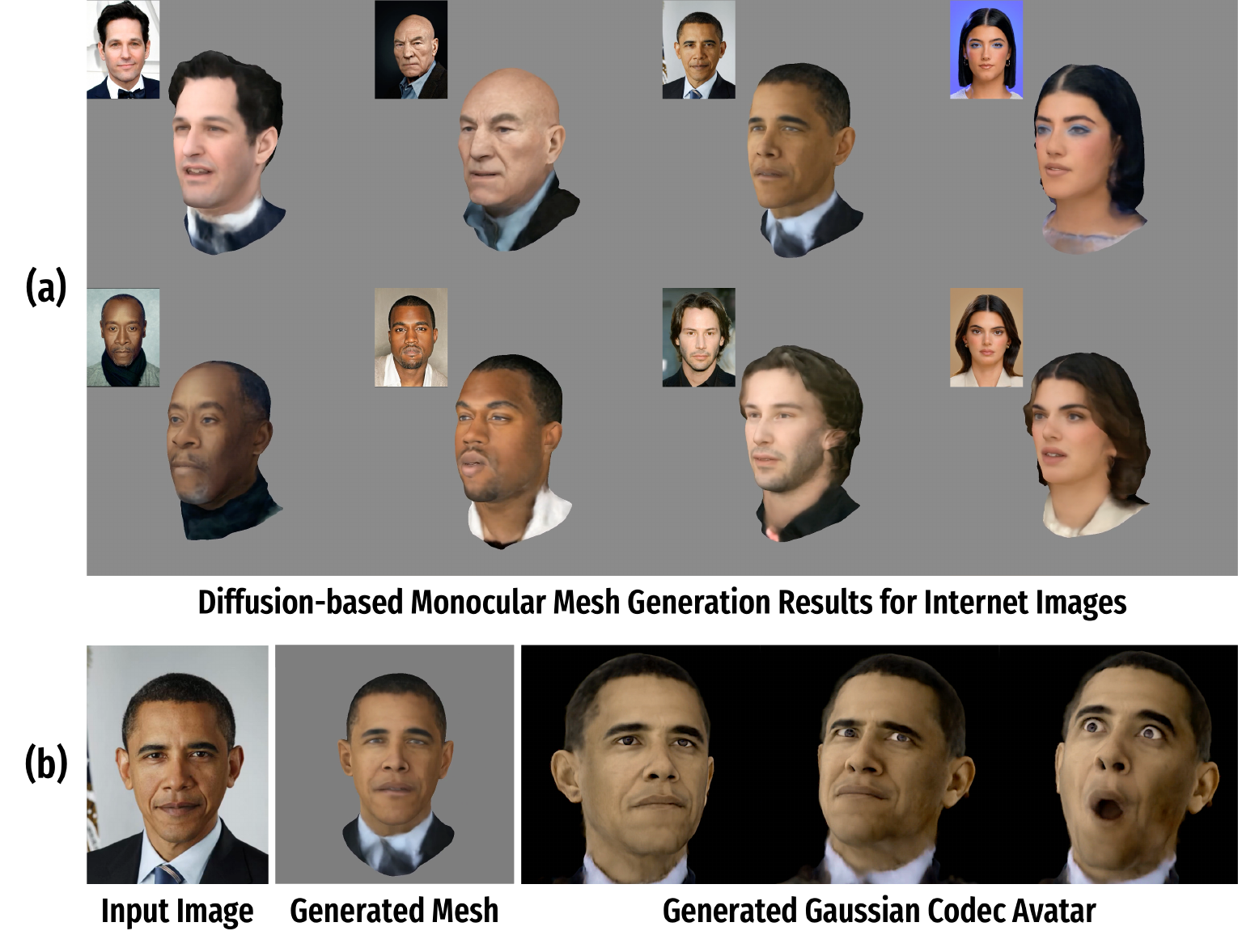}\vspace{-3mm}
    \caption{{Avatar generation result for in-the-wild internet image.}}\vspace{-3mm}
    \label{fig:supp_internet}
\end{figure}

In \Fref{fig:supp_internet}-\colorref{a}, we visualize the generated meshes from our proposed diffusion-based mesh texture \& geometry UV map generation, for the portrait images from the internet.
Although our diffusion model for mesh generation has been trained on 1) a multi-view dome-captured dataset and 2) an iPhone-captured dataset, it generalizes to diverse facial attributes, \eg, make-up, hairstyles, and clothing from in-the-wild portrait images.
We postulate that this generalization capability stems from the model's large-scale human-centric pre-training, which we will detail later in \Sref{sec:supp_ldm}.
The generated meshes are then queried to the Universal Prior Model (UPM), decoded into drivable 3D Gaussian Codec Avatars for real-world telepresence applications (\Fref{fig:supp_internet}-\colorref{b}).

\begin{figure}[t]
    \centering
    \includegraphics[width=1.0\linewidth]{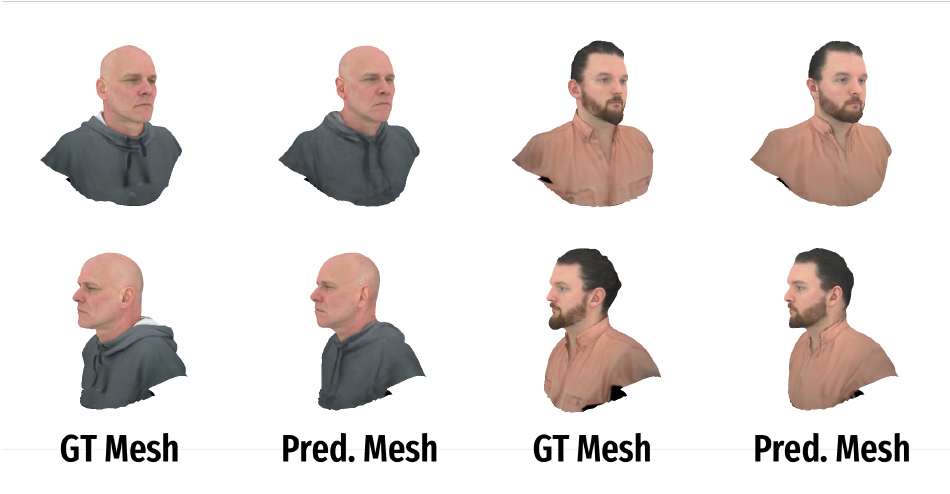}\vspace{-3mm}
    \caption{{\ours~meshes compared with GT meshes.}\vspace{-3mm}
    }
    \label{fig:supp_side}
\end{figure}

In \Fref{fig:supp_side}, we visualize the comparison between \ours-generated 3D meshes and the ground-truth meshes for the unseen test identities. 
From the results, the \ours-generated meshes closely resemble the ground-truth meshes with vivid texture and detailed geometries. Also, note that our generated meshes do not suffer from multi-view inconsistencies, as 
our diffusion-based mesh generation works as UV in-/out-painting, \ie, \ours~generates holistic UV mesh texture and geometry with a single diffusion inference process.

\begin{figure}[t]
    \centering
    \includegraphics[width=1.0\linewidth]{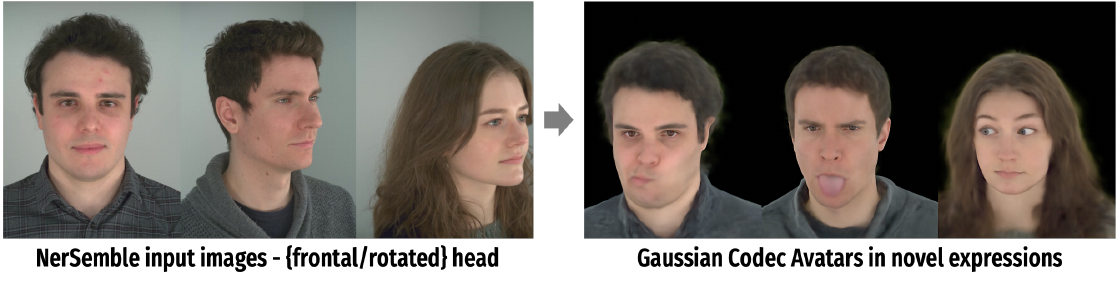}\vspace{-3mm}
    \caption{{Avatar results for NerSemble \{frontal / rotated\} images.}}\vspace{-3mm}
    \label{fig:supp_nersemble}
\end{figure}

In \Fref{fig:supp_nersemble}, we show the Codec Avatar generated from the NerSemble~\cite{kirschstein2023nersemble} identities. These identities are held out, \ie, none of the \ours~modules have seen them during training.
\ours~generalizes well to these held-out unseen identities. Notably, \ours~robustly generates avatars even from input images with oblique head views. 
Although we haven not explicitly designed techniques for these cases, our mesh generation scheme, \ie, UV in-/out-painting conditioned on partially observed visual cues (texture, normal, 3D vertex), helps \ours~generalize to such side-view portrait cases.

\section{Details of \ours~Pipeline}

\subsection{Fine-tuned Sapiens for UV, Normal and Vertex Coordinates Prediction}
Sapiens~\cite{khirodkar2024sapiens} is a human-centric vision foundation model that is pre-trained on large-scale in-the-wild datasets with the masked autoencoder (MAE) task~\cite{he2022maskedautoencoder}.
After pre-training, it can be fine-tuned to perform human-centric perception tasks, such as segmentation or depth/normal estimation.
For our pipeline, we fine-tune the Sapiens models 
for predicting
per-pixel UV coordinates, 
vertex coordinates, 
and normals from a single portrait image. 

\paragraph{Architecture}
We largely follow the architecture of the pre-trained Sapiens-1B model.
For UV coordinates and vertex coordinates prediction, we use the weights of the Sapiens-1B image encoder and add a decoder similar to that of Sapiens-1B (depth).
For normal prediction, we directly start from Sapiens-1B (normal).
We jointly fine-tune the image encoder and the task-specific decoders using a smaller learning rate for the image encoder.

\paragraph{Dataset}
We use an internal iPhone capture dataset containing quarter-body videos of approximately 12,000 identities. 
All frames in the dataset are tracked and annotated using 3D mesh and texture. 
We rasterized UV coordinates, vertex coordinates, and normals into image space to prepare the annotations. 
For vertex coordinates, we adjust the head pose so that the mesh consistently faces forward.

\paragraph{Training}
During training, we sample frames using pre-computed per-frame importance weights to ensure diverse head poses and geometric shapes. 
The frames are augmented with random cropping, scaling, and photometric distortions. 
For UV and vertex coordinates, we use the L1 loss, and for normals, we use cosine similarity loss. 
We use 512 NVIDIA A100 GPUs for 12 hours to train the model for each task.

\subsection{Latent Diffusion Model}
\label{sec:supp_ldm}
In \ours, the core module is the latent diffusion model that generates the complete head texture and geometry in UV maps, given the partial UV observations obtained from the fine-tuned Sapiens models.

\paragraph{Dataset}
We use the UV texture map (${\bT}$) and geometry map (${\bG}$), where ${\bT}, {\bG}\in\mathbb{R}^{H\times W\times 3}$, for training the diffusion model. 
We set $H=W=512$.
The UV texture has a pixel value range similar to that of RGB images. In contrast, the UV geometry maps contain an unbalanced value range across channels, caused by coordinate values from human meshes (high $y$ channel values due to human height).
Thus, we pre-compute the mean and standard deviation of the meshes in our dataset and normalize all the geometry assets.

\paragraph{Architecture} 
The design of our latent diffusion model follows the Diffusion Transformer (DiT)~\cite{peebles2023dit} and Pippo~\cite{kant2025pippo}.
We use the pre-trained SDXL VAE~\cite{podell2024sdxl} and perform 
8$\times$ compression of UV texture and geometry maps, resulting in $32{\times}32{\times}16$ dimension for the latent codes. 
We then patchify the latent codes (of the UV texture and geometry maps) using a linear layer with a patch size of 2. 
We use a fixed sinusoidal 2D positional encoding for the latent patches.

The conditioning data for our diffusion model are the partial UV maps obtained from the Sapiens models and the CLIP image embedding of the reference image.
We follow the pixel-aligned control method, ControlMLP, proposed in Pippo~\cite{kant2025pippo} to condition the model with partial UV maps and generate UV maps.
Also, we inject the CLIP image embedding along with the diffusion timestep embedding in the form of scale, shift, and gate modulation, similar to Stable Diffusion 3~\cite{esser2024scaling}. 
We stack 28 DiT+ControlMLP blocks, and the total number of learnable parameters in the diffusion model amounts to 2B parameters.

\paragraph{Training}
For training our diffusion model, we first conduct image-only pre-training using a large-scale human-centric dataset, following~\cite{kant2025pippo}. For pre-training details, please refer to Pippo~\cite{kant2025pippo}.
Then, we fine-tune the model with pairs of \{portrait image, UV texture/geometry maps\}, via Eq.~(\colorref{1}) in the main paper.
We train the diffusion model for 50K steps with an effective batch size of 128 on 64 NVIDIA A100 GPUs, which takes about 2 days to converge. 

\paragraph{Sampling}
When sampling from the trained diffusion model, we perform 50 steps of flow estimation and updates. 
In a single step, we perform two DiT forward operations by changing the domain switcher $\bd$, that decides which domain (texture or geometry) to predict the flow field for (see Sec.~\colorref{3.1}).
The total sampling time takes about 4 seconds.

\subsection{Feed-forward UV Refinement Net}
For the feed-forward UV refinement network,  
we follow the architecture of \texttt{UNet2DConditionModel} from Diffusers~\cite{patrick2022diffusers}.
As detailed in Sec.~\colorref{3.2}, the input to the UNet is the generated UV texture and geometry maps from the diffusion model. 
The condition to the UNet is the rich image features extracted from the reference image and the mesh rendering.
We use the pre-trained Sapiens ViT encoder as the feature extraction module.
In Eq.~(\colorref{2}) of the main paper, we empirically set $\lambda_\text{pho}{=}2.0$, $\lambda_\text{mask}{=}0.5$, $\lambda_\text{kpts}{=}0.01$, and $\lambda_\text{reg}{=}1.0$.
We train the UV refinement network for 50K steps with an effective batch size of 128 on 32 NVIDIA A100 GPUs, which takes about 2 days to converge.

\subsection{Universal Prior Model}
The Universal Prior Model (UPM) serves as the decoder module to convert the generated UV texture and geometry into the detailed and drivable 3D Gaussian avatar.
We follow the UPM architectures of~\cite{chen2022ipica,li2024uravatar} with several modifications.

First, we broaden the universal corpus of the UPM training dataset, by using the video frames of 1,927 identities (was 255 in \cite{chen2022ipica}, 345 in \cite{li2024uravatar}), captured from 160 multi-view calibrated cameras.
Also, we change the linear color space photometric loss for training the UPM (Eq.~(7) from Chen~\etal~\cite{chen2022ipica}) to the RGB space, using the pre-computed color correction matrix.
This is for the compatibility between the generated UV texture maps and UPM, as \ours~generates RGB space texture maps.
We use 128 NVIDIA A100 GPUs to train the UPM, which takes about 3 weeks to converge.

\section{Broader Impacts \& Ethical Considerations}

\paragraph{Societal Impact}
The primary goal of \ours~is to enabling accessible high-fidelity avatar synthesis for applications in telepresence, mixed reality, and we recognize the potential risks associated with misuse. 
To mitigate these risks, we advocate for the community's ongoing efforts 
in avatar fingerprinting~\cite{prashnani2024avatar} and digital media forensics~\cite{roessler2018faceforensics} to support the detection of synthetic media.

\paragraph{Dataset Disclosure}
We disclose that the collection and use of all human datasets have been conducted in strict accordance with ethical guidelines.
We have obtained informed consent from the subjects involved in the data collection.